\documentclass[letterpaper, 10 pt, conference]{ieeeconf}  

\IEEEoverridecommandlockouts                              

\usepackage{graphics} 
\usepackage{graphicx,subfigure}
\usepackage{epsfig} 
\usepackage{amsmath} 
\usepackage{amssymb}  
\usepackage{multirow} 
\usepackage{algorithm}
\usepackage{algorithmic}
\usepackage{stfloats}
\usepackage{booktabs}
\usepackage{xcolor}
\usepackage{cite}
\usepackage{hyperref}

\usepackage{caption}

\title{\LARGE \bf
Hierarchical Trajectory Planning for Autonomous Driving \\in Low-speed Driving Scenarios Based on RRT and Optimization
}

\author{Yuying Chen$^{1}$, Haoyang Ye$^{1}$, Ming Liu$^{1}$
\thanks{$^{1}$Yuying Chen, Haoyang Ye and Ming Liu are with Department of Electronic and Computer Engineering, The Hong Kong University of Science and Technology. 
        {\tt\small ychenco@ust.hk}, {\tt\small hy.ye@ust.hk}, {\tt\small eelium@ust.hk}}%
}

\begin{document}

\maketitle
\thispagestyle{empty}
\pagestyle{empty}

\begin{abstract}
Though great effort has been put into the study of path planning on urban roads and highways, few works have studied the driving strategy and trajectory planning in low-speed driving scenarios, e.g., driving on a university campus or driving through a housing or industrial estate. 
The study of planning in these scenarios is crucial as these environments often cover the first or the last one kilometer of a daily travel or logistic system. Additionally, it is essential to treat these scenarios differently as, in most cases, the driving environment is narrow, dynamic, and rich with obstacles, which also causes the planning in such environments to continue to be a challenging task. This paper proposes a hierarchical planning approach that separates the path planning and the temporal planning. A path that satisfies the kinematic constraints is generated through a modified bidirectional rapidly exploring random tree (bi-RRT) approach. Following that, the timestamp of each node of the path is optimized through sequential quadratic programming (SQP) with the feasible searching bounds defined by safe intervals (SIs). Simulations and real tests in different driving scenarios prove the effectiveness of this method.
\end{abstract}


\begin{figure*}[!htb]
  \centering
    {\includegraphics[width=1.6\columnwidth]{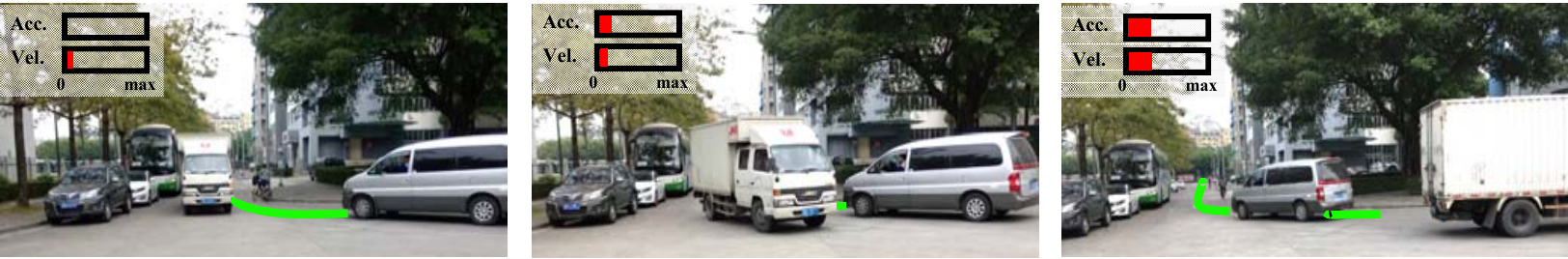} 
  \caption{A typical scenario we explore in this paper. The gray van tries to turn right. Seeing the white truck, it decelerates and waits at the junction. After the white truck passes by, it accelerates and follows the planned path. The green curve shows the planned path and the upper left insets show the velocity and acceleration of the gray van at different moments.
  \label{fig:first}} }
\end{figure*}


\begin{figure*}[!htb]
  \centering
    {\includegraphics[width=1.6\columnwidth]{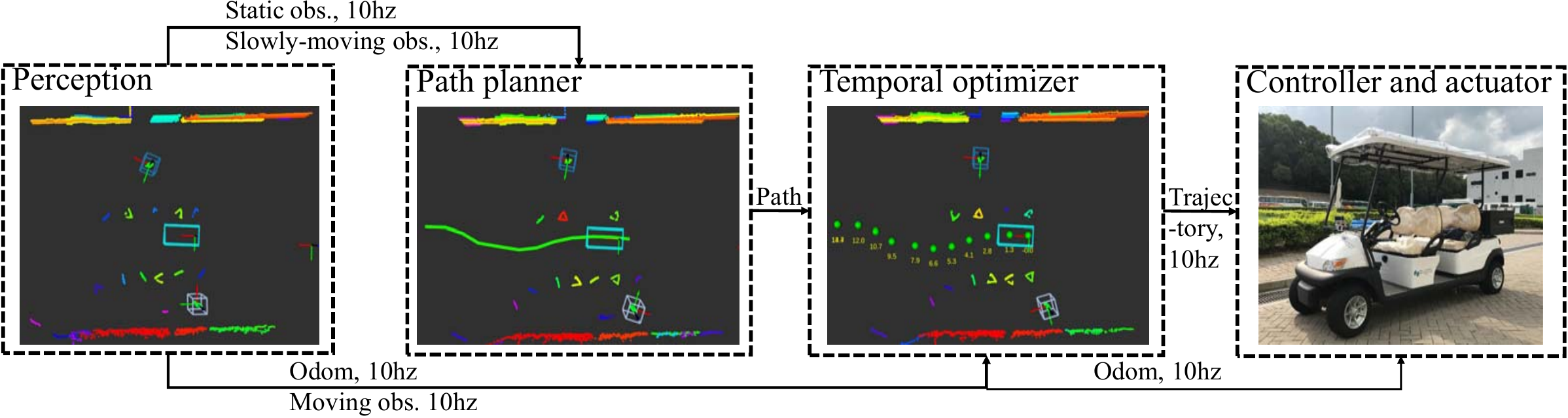} 
  \caption{Functional block diagram of the navigation system. The perception module provides the information of obstacles and the robot. The trajectory planner generates the collision-free path (green curve) and optimizes the timestamp (yellow text) of the nodes (green dots) along the path. Given the trajectory, the controller ensures the robot to execute the plan exactly. 
  \label{fig:function}} }
\end{figure*}


\section{INTRODUCTION}

A boom in the study of planning for car-like robots has appeared since the DARPA Grand Challenge and Urban Challenge. 
This is helping to propel the advancement of autonomous driving, which is anticipated to improve the safety and efficiency of the transportation system, and offer outstanding convenience to users. 
While most of the current work concentrates on driving scenarios in which cars run along highways or urban roads, where the traversable area is confined by structured lane markings and barriers, few studies pay attention to obstacle-rich or narrow environments, or driving through additional moving obstacles, e.g., pedestrians and other vehicles. 
Despite this, trajectory planning in these scenarios is of great research and development value. 
A typical scenario is shown in Fig.\ref{fig:first}, which occurs in an industrial estate. 
The complex environment and the various road participants challenge the robustness of the planner and its adaptivity to the dynamics, which continue to be important issues. 
Furthermore, from the perspective of popularizing autonomous driving technology, it is fundamental to satisfy the transportation requirement of daily life, which usually includes a trip from a residential district to the work place. 
Other low-speed driving scenarios, including driving on a university campus, driving in an industrial estate, or parking in an open parking lot are also common.

The difficulties in planning can be viewed from two aspects: the cluttered environment, and the dynamics of the complex scenario. 
Arising from the navigation solution in the static environment, the most common scheme is path generation and following. 
Taking no account of the performance, it is workable in most low-speed cases as the path planners handle the cluttered environment and help to avoid static obstacles efficiently. 
An extra obstacle avoidance module or replanning helps the car to change its route and keep its distance from obstacles. 
However, it requires a relatively spacious area and may leave the car trapped in the scenarios we focus on in this paper. 

For narrow or cluttered environments, it is less convenient to passively avoid dynamic obstacles. 
Therefore, adding a time dimension to the planning is essential to cope with the dynamic environment. 
Additionally, planning should meet the real-time requirement for avoiding collisions. 
Under these conditions, to enable vehicles to travel through a complex dynamic environment safely and efficiently, we propose a real-time hierarchical planning approach, taking advantage of both the efficiency of the sampling based methods for path planning and the flexibility of optimization approaches for time scheduling. 
A smooth path that satisfies the kinematic constraints of the car is generated by the modified bidirectional rapidly exploring random tree (bi-RRT) method. 
Then the safe intervals (SIs) along the path are estimated and confine the searching time region for further optimization by sequential quadratic programming (SQP). 
The temporal planning of each pose is optimized in the sense of a customized cost function. 
The motions of dynamic obstacles are tracked by multiple Kalman filters. Though they are predicted to move with uniform speeds, it is practicable with the update rate of 10 HZ in the low speed scenarios. 
Overall, this work makes the following contributions:
\begin{itemize}
\item We propose a hierarchical scheme for trajectory planning that by combining the bi-RRT path planning and temporal optimization, can perform real-time and safe driving of a car in dynamic and cluttered environment.
\item We propose a revised bi-RRT planner for fast searching that also suits car dynamics. 
\item We propose a temporal planning method based on the idea of SIs, and build a complete system including updating and emergency/time-out recovery.
\item Through extensive simulation and real tests, we show the robustness and good performance of our system in various complex environments. 
\end{itemize}

\section{Related work}
\subsection{Path planning}
Many autonomous vehicles have been demonstrated successfully driving through a specific route \cite{urmson2008autonomous,montemerlo2008junior,ziegler2014making}. Behind this, various methods have been proposed to solve the planning problem with different strengths and weaknesses. 
Graph-search-based planners like Hybrid A* \cite{montemerlo2008junior} and Spatio-temporal lattice \cite{ziegler2009spatiotemporal} search the state space represented by an occupancy grid or state lattices and yield a global optimum. However, with the dimension of space increasing, the time cost and memory consumption increase increase exponentially. Furthermore, it is critical to build the graph for unconventional road environments.    
Methods based on piecewise curves like polynomials \cite{bacha2008odin}  are suitable for a structured environment, but typically rely on prestored sampled trajectory candidates. This generate-and-evaluate scheme reduces the computational cost, but lacks flexibility and long-horizon prediction. 
Similar to the spatio-temporal lattice method, optimization methods like that in \cite{ziegler2014trajectory} can be used for spatial temporal planning. These methods minimize the customized cost so as to bring a well-rounded solution that can be collision-free and have both low cost and high comfort. However, these methods suffer from long computation time.
In contrast, sampling based methods like RRT \cite{kuwata2009real} randomly sample the configuration space and construct the tree incrementally, in which case a fast solution can be provided. It avoids the discretization of the state space and requires no prior information of the environment. Thus, it ensures both flexibility in generating the path and applying it to various environments, regardless of the structures.


\subsection{Dynamic obstacle handling}
Dynamic obstacles composed of pedestrians and vehicles bring great uncertainty to the driving environment. With the exception of several learning-based methods that make the robots avoid pedestrians in a human manner with only raw inputs \cite{tai2017socially}, the strategy of modelling the motion of obstacles first and then planning dominates. 

One common practice is to assume that obstacles move at a known and constant velocity (CV) within a short time horizon. However, it is crucial to consider the uncertainty of observations of obstacles and their real motion. The Bayesian occupancy filtering proposed in \cite{coue2006bayesian} utilizes a probabilistic grid representation of the dynamic surroundings and is further used for danger estimation as well as collision avoidance. In \cite{kushleyev2009time}, by assuming constant controls of obstacles in the near future, the trajectories of the obstacles are modelled as a series of Gaussian distributions by the prediction step of the extended Kalman filter. Then possibilities of collision along the time-bounded lattice are calculated and regarded as a determinant of the final trajectory.  Similarly, in this paper, the moving obstacles are tracked with multiple Kalman filters; however, their motions directly influence the SIs of the nodes along the path and thus affect robot movements.

 
With CV assumption, methods based on the concept of velocity obstacle (VO) \cite{fiorini1998motion} define either collision cones or half planes \cite{berg2011reciprocal} on the velocity space and provide solutions of collision-free velocities. Extensions, including bicycle reciprocal collision avoidance \cite{alonso2012reciprocal}, further make these methods applicable for cars. These agent-based models perform well with homogeneous moving objects but can easily get the robot stuck if the environment becomes cluttered. As the free space is usually limited compared to the size of the robot, instead of taking any collision avoidance action or frequently replanning, other methods try to incorporate the dynamic information into planning at the very beginning. In \cite{phillips2011sipp}, the 4D search space (x,y,$\theta$,time) is constructed with states defined by configurations and SI. By applying A*, it demonstrates real-time feasible planning in dynamic environments. However, it still suffers from the aforementioned limitations of A*. Instead of 4D planning, speed profile planning \cite{liu2017speed} along a preplanned path is proposed to avoid collision, which can obtain a fast solution. In \cite{lim2018hierarchical}, speed planning is performed for each path candidate and trajectory with the least cost is selected.
Inspired by the ideas of safe interval and speed profile planning, we propose a hierarchical planning method that combines sampling-based path planning and temporal optimization to get real-time performance.

\section{The planning algorithm}
\subsection{Nomenclature} 
The notations used in this paper are declared in Table \ref{tab:notation}. It lists the symbols that appear in the algorithms or equations that are not clearly elucidated.
\begin{table}[htbp]
  \centering
 \caption{The notations in this paper.\label{tab:notation}}
 \begin{tabular}{lcl}
  \toprule
  Symbols & Meaning\\
  \midrule
$P_{start}$ & start pose of the car\\
$P_{t}, v_t$ & pose and velocity of moving obstacles\\
$a, b, c$ & coefficients of the curvature representation\\
$s_f$ & total length of the curve\\
$s$ & accumulative arc length along the curve, $s\in{[0,s_f]}$\\
$x_{rand}$ & randomly generated point\\
$n_{rst}$ & nearest node\\
$T_{f}$ & forward growing tree grown from the start\\
$T_{b}$ & backward growing tree grown from the goal\\
$p$& random value in $[0,1]$ with a uniform distribution\\
$d_{th}$ &  threshold distance between two trees\\
$p_{th}$&  probability of using GMM data sampling\\
$r$ & distance between the node and $x_{rand}$\\
$\beta$ & azimuth angle of $x_{rand}$ in the coordinate of $n_{rst}$\\
$P_i$ & pose of $i$th node, represented by $(x,y,\theta)$\\
$t_i$ & timestamp of $i$th node \\
$v_i$, $a_i$& velocity and acceleration of $i$th node\\
$start_{ij}$ & start timestamp of the $j$th SI of node $i$\\
$k_{i}$ & number of safe intervals of node $i$ \\
  \bottomrule
 \end{tabular}
\end{table}


\subsection{Framework}
\begin{figure}[!ht]
  \centering
    {\includegraphics[height=0.7\columnwidth]{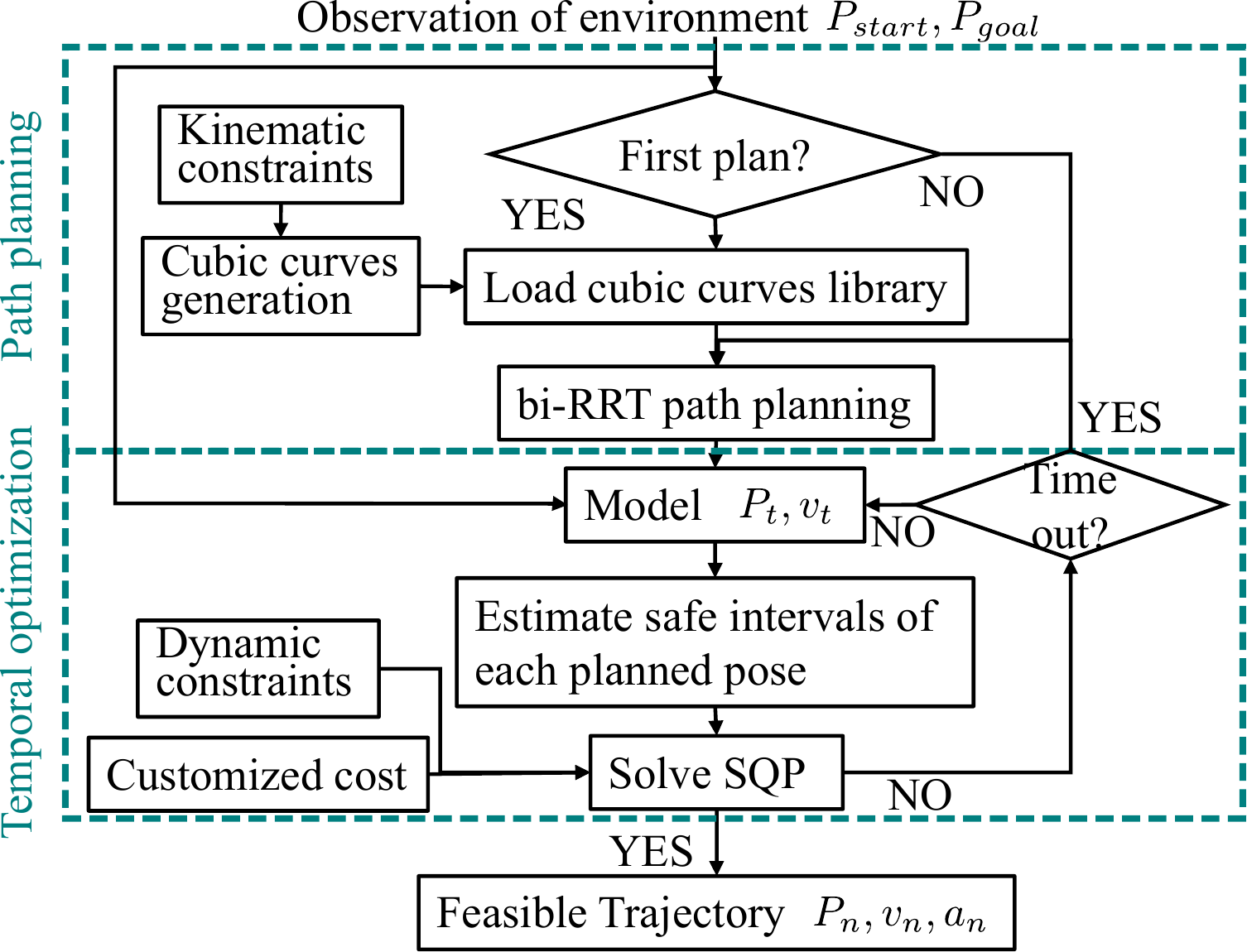} 
  \caption{Flow chart of the proposed trajectory planning method
  \label{fig:framework}} }
  \vspace{-1em}
\end{figure}

The framework of the proposed method is shown in Fig.\ref{fig:function}. Taking both the input and output into account four modules are presented. The algorithm relies on a perception module to provide the information of the obstacles and the car itself. Additionally, a control module translates the trajectory into control commands. More details on the path planner and temporal optimizer are shown in Fig.\ref{fig:framework}. For the path planning part, we first generate a curve library off-line and load it into the memory for the first plan. After receiving the sensor data and command of the goal, the modified bi-RRT is executed for path planning. To tackle dynamic obstacles in the environment, firstly the states of the obstacles will be modelled, which also provides information for the SI estimation of each pose. After optimizing the customized cost given the dynamic constraints and searching the boundaries defined by the SIs, suggested timestamps of each pose will be generated along with the feasible trajectory. The observations of the environment are updated at all times. If the optimizer fails to find a solution, a timer will be started and the state of the obstacles will be estimated again. If the time is up, the planner will replan the path. So, in most cases, replanning is not happening except when moving obstacles stop and block the way.    
 
\subsection{Collision checking}
\begin{figure}[!htb]
  \centering
{\includegraphics[width=0.75\columnwidth]{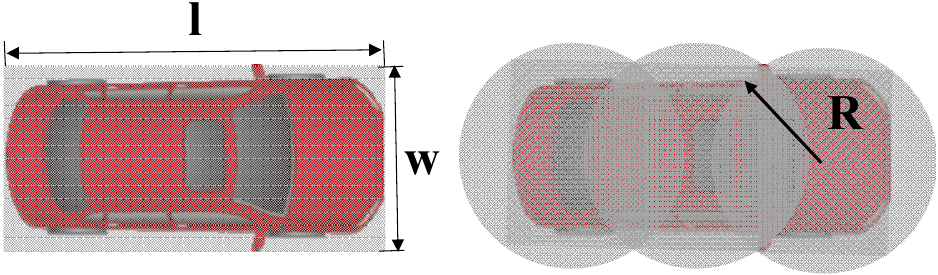} 
  \caption{Car modelling for fast collision checking
  \label{fig:car1}} }
  \vspace{-0.5em}
\end{figure}

Collision checking is important for both the tree growing in path planning and SI estimation in temporal optimization. The most typical 2D representation is the rectangle, which is a perfect match for most types of vehicles and is conventional in the perception process (object detection and tracking). However, due to the anisotropy, it is not convenient for collision checking. In this paper, a representation of three overlapped circles that covers the footprint of the car is applied,  as shown in Fig.\ref{fig:car1}. For moving objects like humans that have larger aspect ratios (w/l), the three-circles representation will induce more redundant area than the footprint; therefore, these obstacles are represented by one circle. The radius of the circles in each case is shown in Eq.\ref{eq:radius}.
\begin{equation}
\label{eq:radius}
R=\left\{  
             \begin{array}{lr}  
             \sqrt{(l^{2}+w^{2})/4}, & l/w < 1.3 \\  
             \sqrt{(l^{2}+9w^{2})/36}, & l/w\geq1.3    
             \end{array}  
\right. 
\end{equation}


\subsection{Path planning}
We propose the modified bi-RRT method for path planning, which combines the speed and flexibility of RRT and also satisfies the kinematic constraints by replacing the line segments between the nodes with curves in the lookup table. In addition, the bidirectionally grown two trees stretch from the start and goal respectively, which saves time, especially in a complex environment. 

\begin{figure}[!htb]
  \centering
     \subfigure[curve representation]{\includegraphics[height=0.22\columnwidth]{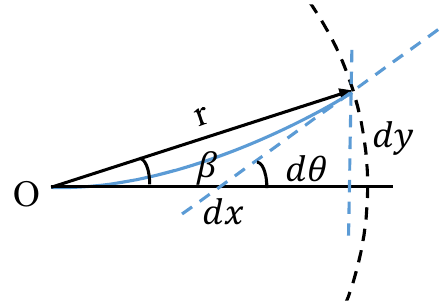}
    \label{fig:curve1}}
      \subfigure[the curve samples]{\includegraphics[height=0.22\columnwidth]{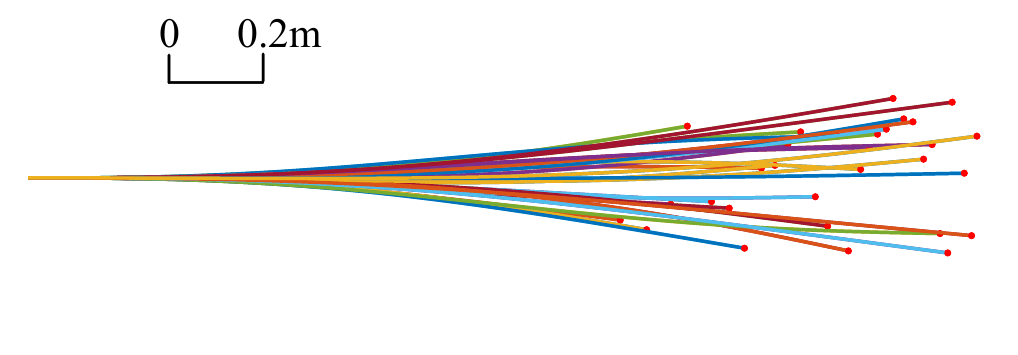}
    \label{fig:curve2}} 
  \caption{(a) Representation of a cubic-curvature curve stored in the lookup table. (b) Curve samples in the curve library.}
  \vspace{-0.5em}
\end{figure}

The cubic-curvature curves with limited curvature are generated based on the method described in \cite{nagy2001trajectory}. As shown in Eq.\ref{eq:curvature}, the curvature of the curve is a third order polynomial in arc length. Given the curvature, the function of the changes in heading angle $\theta$ (Eq.\ref{eq:deltatheta}) and position (Eq.\ref{eq:deltax} and Eq.\ref{eq:deltay}) can be represented. The final curve can be represented by four parameters $[a,b,c,s_f]$ and we sample the end pose of the curve on the circle with radius $r$ (shown in Fig.\ref{fig:curve1}). The radius $r$ and the angle $\beta$ are recorded for fast lookup. So the final representation of the curve in the lookup table is $[r,a,b,c,s_f,dx,dy,d\theta,\beta]$. The curve samples in the curve library are shown in Fig.\ref{fig:curve2}.

\begin{align} 
&\kappa(s) = \kappa_0 +as+bs^2+cs^3,  \kappa_0,a,b,c\in \mathbb{R}\label{eq:curvature}\\
& d\theta(s) =\int_0^s \kappa(l)dl = \kappa_0s +as^2/2+bs^3/3+cs^4/4 \label{eq:deltatheta}\\
&d x(s) = \int_0^s cos(\theta(l))dl \label{eq:deltax}\\
&d y(s) = \int_0^s sin(\theta(l))dl \label{eq:deltay}
\end{align}

\begin{figure}[!htb]
  \centering
{\includegraphics[width=0.75\columnwidth]{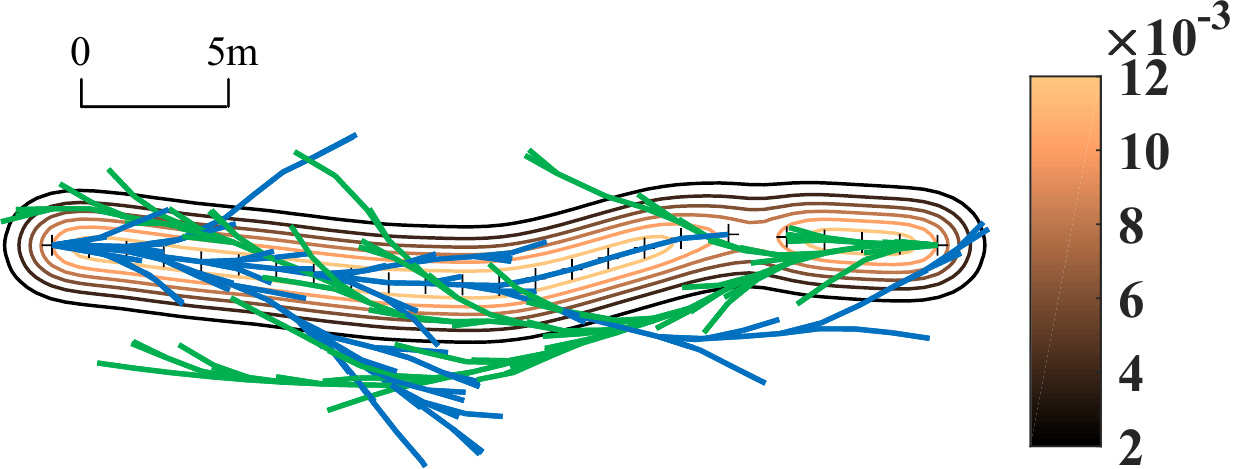} 
  \caption{Gaussian mixture distribution data sampling. The black crosshairs represent the nodes locating at the potentially closest path. The blue and green curves show the forward and backward tree respectively. The contour lines show the probability distribution of the GMM. 
  \label{fig:gmm}} }
  \vspace{-0.5em}
\end{figure}

After loading the lookup table, the bi-RRT method will grow two trees incrementally. The algorithm can be seen in Algorithm \ref{alg:rrt}, which includes sampling a node, extending the tree and connecting two trees. Some modifications have been made for it to be suitable for car-like robots. As shown in  Algorithm \ref{alg:extend}, while extending the tree, we replace the line segment between $x_{new}$ and  $n_{rst}$ with a curve that possesses an end point close to the extending direction, and is chosen to optimally fit the line. In order to prevent repetition of nodes, each curve can only be chosen once for the same node. For trees connection, we use C Feasible SQP\cite{lawrence1994user} to online calculate the connecting cubic-curvature curve if the matched two nodes satisfy several rough conditions like distance and heading changes. As the curves with $\beta\in\{\beta_{min},\beta_{max}\}$ are more likely to be chosen, the trees spread fast. However, it is easy to generate a loop structure and it costs much time to connect the two trees. In  Algorithm \ref{alg:sample}, in order to speed up the connection of the two trees, we introduce a sampling method that generates samples following the distribution of a Gaussian mixture model after they get close enough. As shown in Fig.\ref{fig:gmm}, the centers of the Gaussians are located at the nodes of the potentially shortest path. In the following, our introduced sampling strategy will be referred to as the GMM sampling, and the pure random sampling strategy will be referred to as the Random sampling. 

\begin{algorithm}[!ht]
    \caption{bi-RRT for car-like robots}
    \label{alg:rrt}
    \begin{algorithmic}[0]   
    \STATE{
    Load lookup table}\\
    \FOR {i = 0 to n}
    \STATE{$x_{rand}$ = Sample(i)}\\
    \STATE{$n_{rst}$ = FindNearestNode($x_{rand}$,$T_{f}$)}\\
    \STATE{$n_{new}$ = Extend($n_{rst}$, $x_{rand}$)}\\
    \STATE{Insert($n_{new}$,$T_{f}$)}\\
    \IF {SearchNodesNear($n_{new}$,$T_{b}$,radius)$>$1}
    	\STATE{Connect($T_{f}$,$T_{b}$)}
    \ENDIF
    \STATE{Swap($T_{f}$,$T_{b}$)}
    \ENDFOR\\   
\end{algorithmic}
\end{algorithm}

\begin{algorithm}[!ht]
    \caption{Sample(i)}
    \label{alg:sample}
    \begin{algorithmic}[0]
    \STATE{ $p =$ Rand()}
    \IF {FindNearestDistance($T_{f}$, $T_{b}$)$<d_{th}$ and $p<p_{th}$}
    	\STATE{ return GMMSample($n_{start},n_2,...,n_{goal}$)}
    \ELSE
    	\STATE{ return RandomSample($n_{start}$,$n_{goal}$, eccentricity)}
    \ENDIF     
\end{algorithmic}
\end{algorithm}

\begin{algorithm}[!ht]
    \caption{Extend($n_{rst}$, $x_{rand}$)}
    \label{alg:extend}
    \begin{algorithmic}[0]
    \STATE{$r$ = Distance($n_{rst}$,$x_{rand}$)}
    \STATE{$r = min(max(r,r_{min}), r_{max})$}
    \STATE{Calculate $\beta$ in the coordinate of $n_{rst}$.}
    \STATE{$\beta = min(max(\beta,\beta_{min}), \beta_{max})$}
    \IF {curve with entry ($r$, $\beta$) not visited at $n_{rst}$}
    \STATE{Look up table and get the representation of the curve.}
    \STATE{Return $n_{new}$}
    \ENDIF     
\end{algorithmic}
\end{algorithm}

\subsection{Temporal optimization}
Given a feasible path, temporal optimization can be implemented.
However, it is hard to map the continuous path to a continuous duration of time. The alternate methods are that we either discretize the timeline and find the best locus on the path for each timestamp, or sample the path and find the best timestamp for each locus. For RRT, the generated path composed of nodes and edges is discretized naturally. It is natural to optimize the path-to-timeline mapping by finding the best timestamp for each node.

\begin{figure}[!ht]
  \centering
    {\includegraphics[width=1.0\columnwidth]{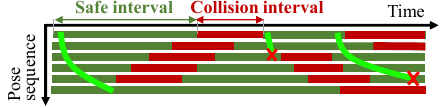} 
  \caption{Demonstration of SIs of the nodes on the planned path, and construction of the layered structure of the SIs for selecting a sequence of intervals for SQP. Each SI is a candidate in a layer. The green curves show the selection of SIs. The red crosses show the early termination of the growing sequence.
  \label{fig:safe}} }
  \vspace{-1em}
\end{figure}

The idea of the SI is explicit. As demonstrated in Fig.\ref{fig:safe}, given the pose of each station, it is the largest time period that ensures no collision, which means that extending the time in either direction would cause the vehicle to have a collision. Correspondingly, the complementary time intervals are the collision intervals. Here, due to the dynamic change of the surrounding environment, the SIs of successional poses show some specific patterns. For example, the line pattern in the middle can be attributed to a moving car. Taking advantage of the SI, if we set the timestamp of every node within its SIs and ensure that the distance between any two contiguous nodes is adequately small, the trajectory is collision-free. The length of all curves is set to be smaller than the size of the car to ensure that the SI is feasible. Additionally, optimizing the timestamp at each node instead of at densely-sampled locations can reduce the computational cost and save time, and, most importantly, still guarantee safety. 

 In this premise, if we define the velocity and acceleration of the car at each node as shown in Eq.\ref{eq:vi} and Eq.\ref{eq:ai}, respectively, the temporal optimization problem can be represented (shown in Eq.\ref{eq:optimization}). The target of time optimization is to find the optimal timestamp of each node or an optimal $s(t)$, where $s$ is the length along the path. To satisfy the collision free requirement, the solution space of each timestamp is limited to its corresponding SIs. As we do not assume a reverse, $s(t)$ should be non-decreasing and the timestamp of successive nodes should be increasing. In addition to the condition of SI, other dynamic constraints, like velocity and acceleration speed limits, should also be satisfied. For the expectation of the optimization, less time and less velocity variation will enhance the driving comfort. So the optimization object function is set to both minimize the time and acceleration speed. Here, the weights $w_t = 1/t_{min}^2$ and $w_a = 1/[(n-2)a_{max}^2]$ are chosen for normalization. The problem cannot be directly solved by SQP as the inequality constraints are disjunct. It can be solved by mixed-integer nonlinear programming (MINLP) or generalized disjunctive programming (GDP) by inducing the integers for decision. However, these processes are time-consuming. 
 
To simplify the searching process for a SIs sequence, we construct a layered structure (shown in Fig.{\ref{fig:safe}). SIs of the same pose sequence are considered to be candidates in a layer. Starting with intervals with the smallest pose sequence, we grow the SI sequence till the last layer. For every valid SI in the $i$th layer (with pose sequence $i$), we select children from the SIs with the pose sequence $i+1$. The criteria is that the two intervals overlap for certain length of time to ensure that the car can pass successfully. Every time a child-parent relationship is established, the SI of the child will be modified to also consider the limit from their parent. Only the SIs being selected are considered to be valid and can go on to find their child. When it comes to the last layer, the SI with the smaller time will be more likely to be picked. And it can go back to the parent to get the SI sequence. 
After getting the proper SI sequences, the final optimized solution is captured by several SQP iterations.
 \begin{align}
 &v_i = \frac{P_i-P_{i-1}}{t_i-t_{i-1}}, i=2,...,n \label{eq:vi}\\
 &a_i = \frac{v_i-v_{i-1}}{t_i - t_{i-1}}, i=3,...,n\label{eq:ai}
 \end{align}
\begin{equation}
\begin{aligned}
& \underset{t_1,t_2,...,t_n}{\text{min}}
& & \underbrace{w_tt_n^2}_{\text{time cost}} +\underbrace{w_a\sum_{i=3}^{n} a_i^2}_{\text{acceleration cost}} \\
& \text{s.t.} & &  t_i \in [start_{ij}, end_{ij}], i = 1,2,....n,j=1,...,k_i\\
& & &  t_i  \le t_{i+1}, i = 1,...,n-1 \\
& & &  v_i  \le v_{max}, i = 2,...,n \\
& & &  a_i \le a_{max}, i = 3,...,n. \\
\end{aligned}
\label{eq:optimization}
\end{equation}

\section{Experimental results}
\subsection{Experimental setup}
The proposed trajectory planner has been tested in both simulation and real environments. We constructed simulation environments in Stage\cite{vaughan2008massively} with different static and dynamic obstacles and implemented repeated experiments by strictly controlling the start and goal of the trajectory and also the states of the dynamic obstacles. The environment in simulations is usually simpler than in the real case, particularly with respect to dynamics. Also, sensors in the real environment receive more noise and this may affect the planning. In the sense of robustness, we tested similar scenarios in the real environment with the golf cart shown in Fig.\ref{fig:function}. It is equipped with one 16-line lidar for both localization and planning. Furthermore, it is comparable in size and has similar kinematic constraints to the car in simulations. However, due to the poor repeatability of the test conditions in the real environment, it is hard to perform quantitative, massive data analysis. As the tests involve interaction between the car and the complex environment, it is not equitable to compare it with planners that can only handle static obstacles or agent based algorithms that mainly focus on the dynamic obstacles. 
We compared the planner with a human driver who tries to reach the goal in the shortest time while maintaining a proper driving style and complying with the same velocity limitations.   

\subsection{Path planning}
To show the effectiveness of the GMM sampling, we compare the GMM sampling and Random sampling method in the same condition. As shown in Fig.\ref{fig:comparison}, the goal point is (20,20,$45^{\circ}$) and start point is (0,0,$0^{\circ}$). The obstacles are 15 randomly generated disks with radius in the range of [0.5,2.0]. Given the same start pose and goal pose as well as the same environment, the sampling points of the GMM sampling are more concentrated on the adjacent region of the final path, while the points of random sampling spread in the space and cause a bigger tree and longer path. As the grown trees are also influenced by the pose difference between the start point and the end point, we test the planner with different sampling methods and different starting poses. 
Fig.\ref{fig:comparison}(b) shows the statistical results of the number of nodes, planning time and path cost with different start angles of 313 runs. It is clear to see the reduction of nodes for the GMM sampling methods. But this reduction of sampling nodes does not necessarily lead to a reduction in time as the method can not reduce the times of optimization, which contributes to a large proportion of the planning time. 
Note that planner with GMM sampling plans shorter path than with Random sampling in all cases. 
GMM sampling reduces the cost when there is a large heading difference between the start and goal. 

As is shown in Fig.\ref{fig:planning}, the planner is also tested in several representative scenes. The first scene is the vehicle turning right. In the second scene, the vehicle is turning left at a sharp corner, and there are several cars parked on the road. The third scene shows the vehicle trying to overtake a slow car to the right. There is also a pedestrian moving in the opposite direction to the left. These three scenes are very common in a residual or an industrial estate, but are quite different to the case of urban roads or highways as they are less structured and more cluttered. 
\begin{figure*}[!ht]
  \centering
    \subfigure[Qualitative analysis]{\includegraphics[height=0.55\columnwidth]{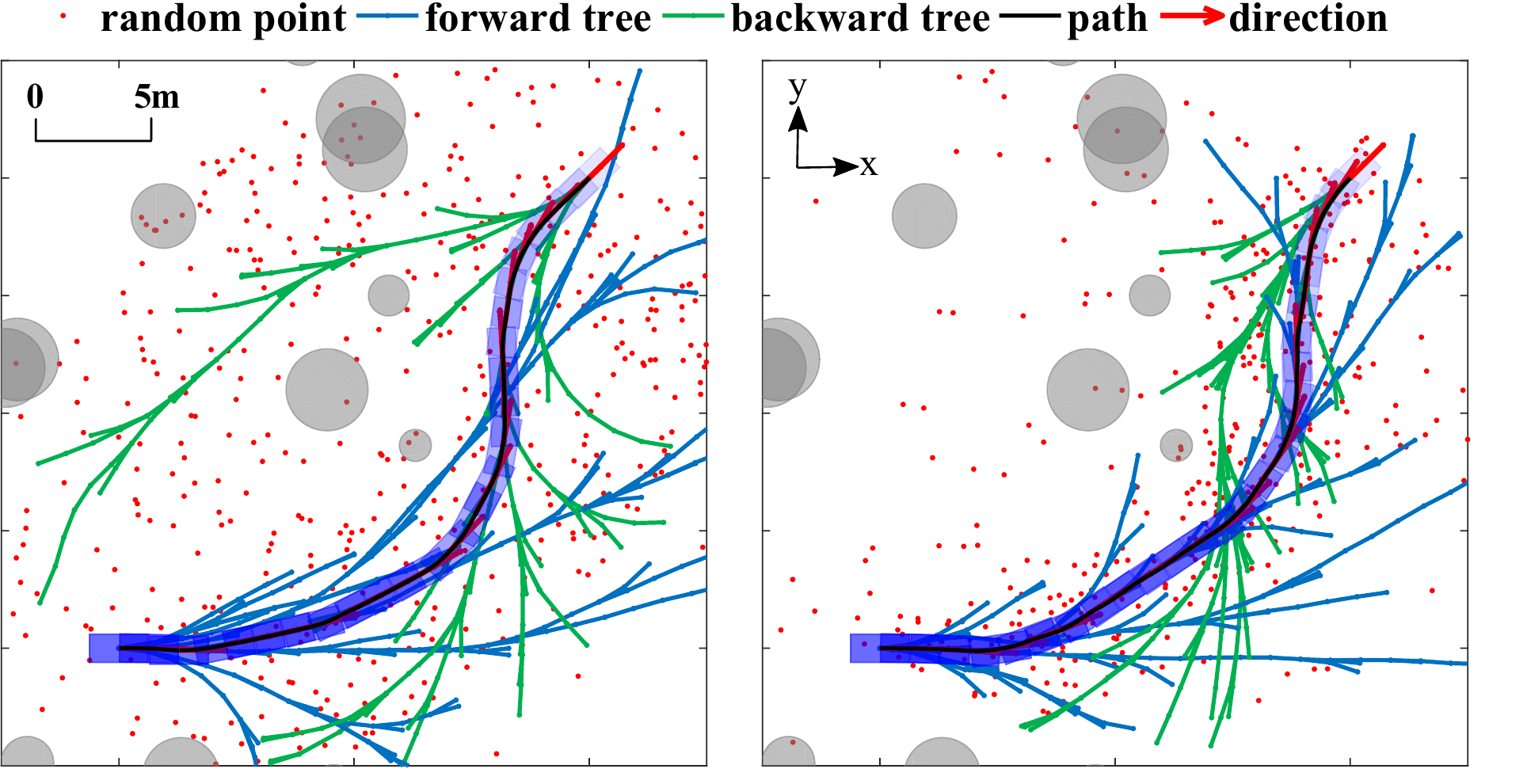}} 
    \subfigure[Quantitative analysis]{\includegraphics[height=0.55\columnwidth]{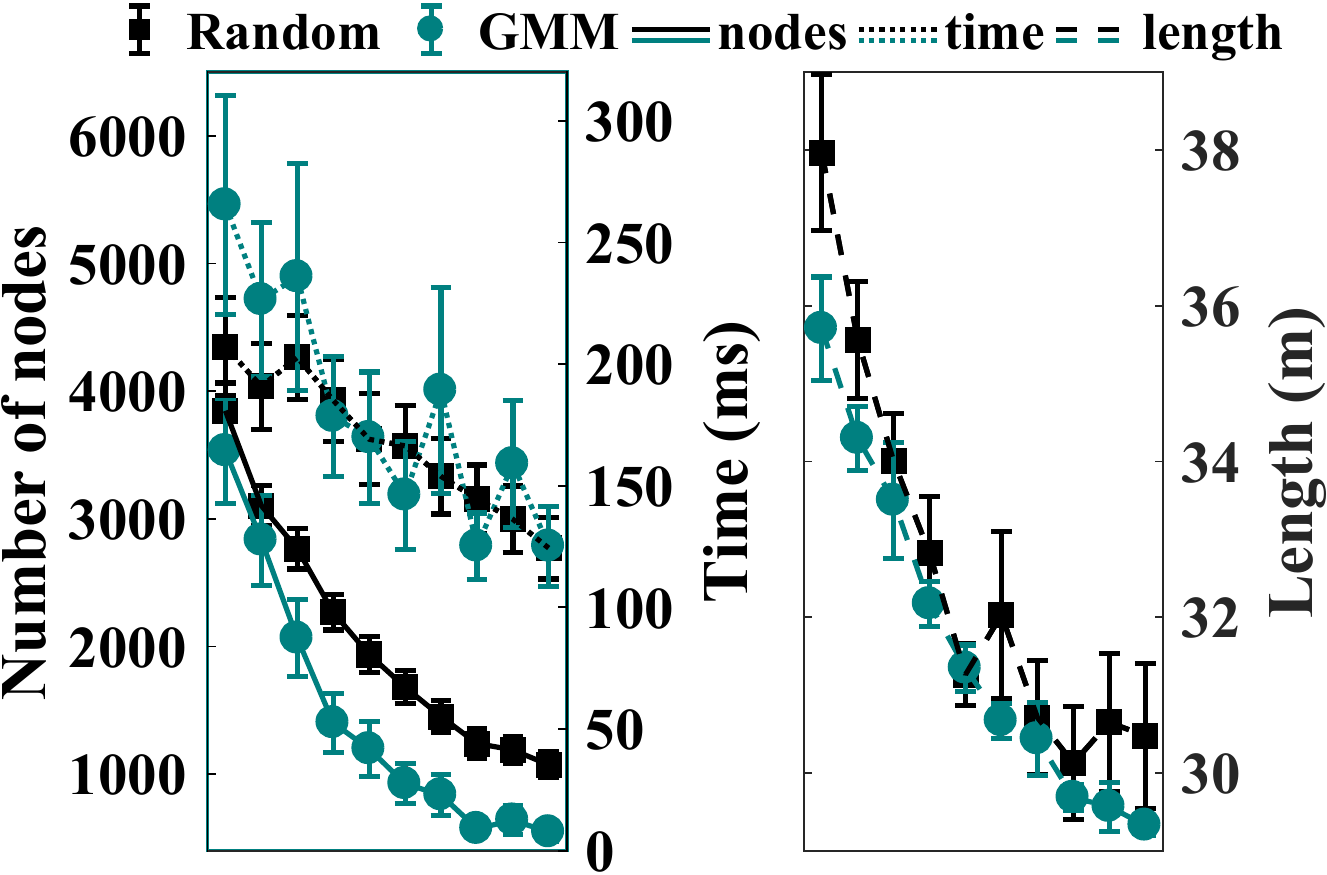}}
  \caption{(a) Qualitative analysis of planners with Random samping only (left) and GMM sampling (right). The gray disks are the randomly generated obstacles. The red dots show the random seeds for tree growth. (b) Quantitative analysis of planners with Random samping only (black) and GMM sampling (blue). The results of two sampling methods over 10 heading differences ([$90^{\circ}, 80^{\circ},70^{\circ},60^{\circ},50^{\circ},40^{\circ},30^{\circ},20^{\circ},10^{\circ},0^{\circ}$]) are shown here, with the markers showing the average value of the trials and the error bars representing the 95\% confidence intervals. \label{fig:comparison}}
  \vspace{-1em}
\end{figure*}


\begin{figure*}[!ht]
  \centering
  
  \subfigure[Turn right]
{\includegraphics[width=0.55\columnwidth]{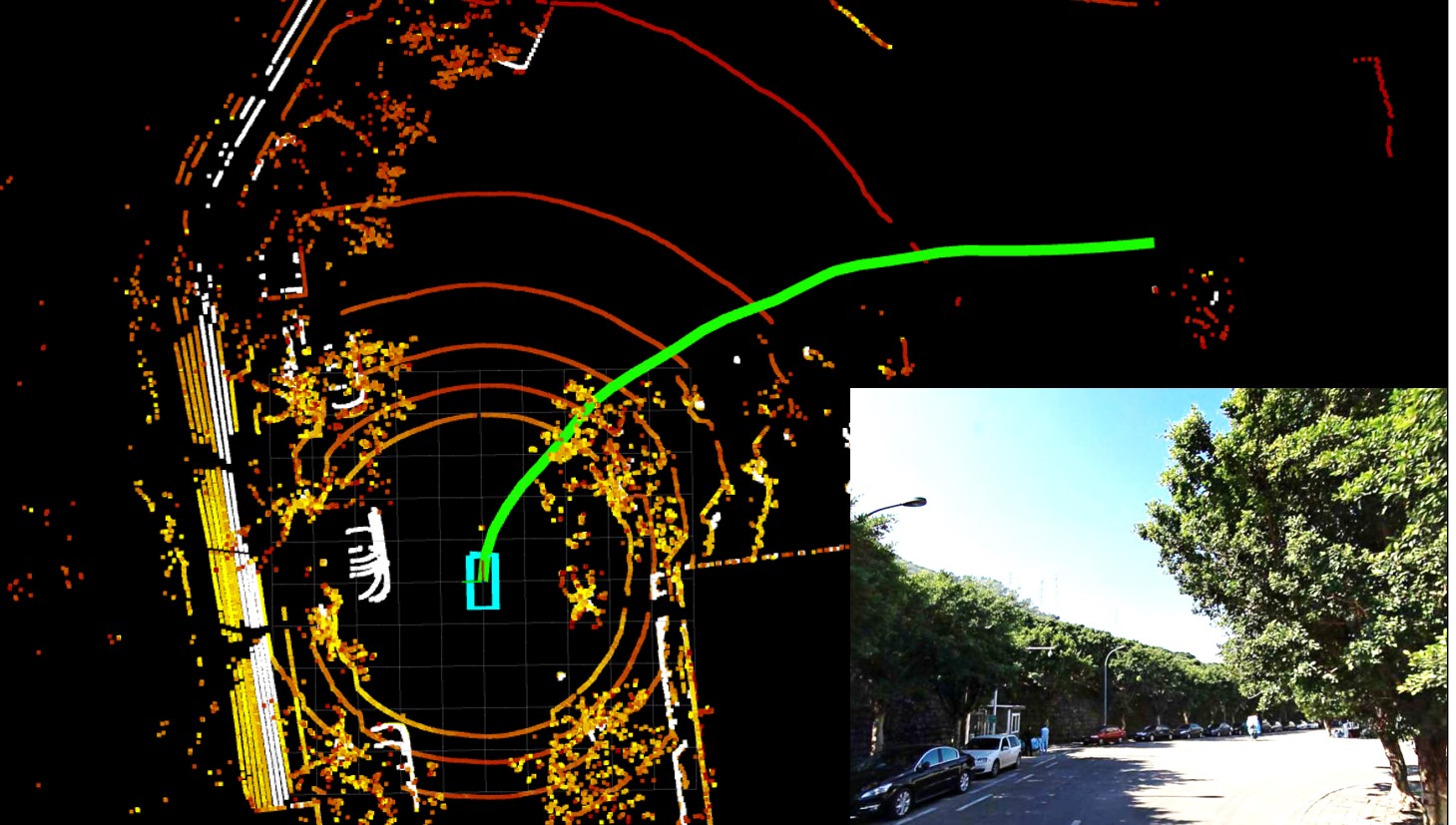}
    } 
     \subfigure[Turn left at a sharp corner]{\includegraphics[width=0.55\columnwidth]{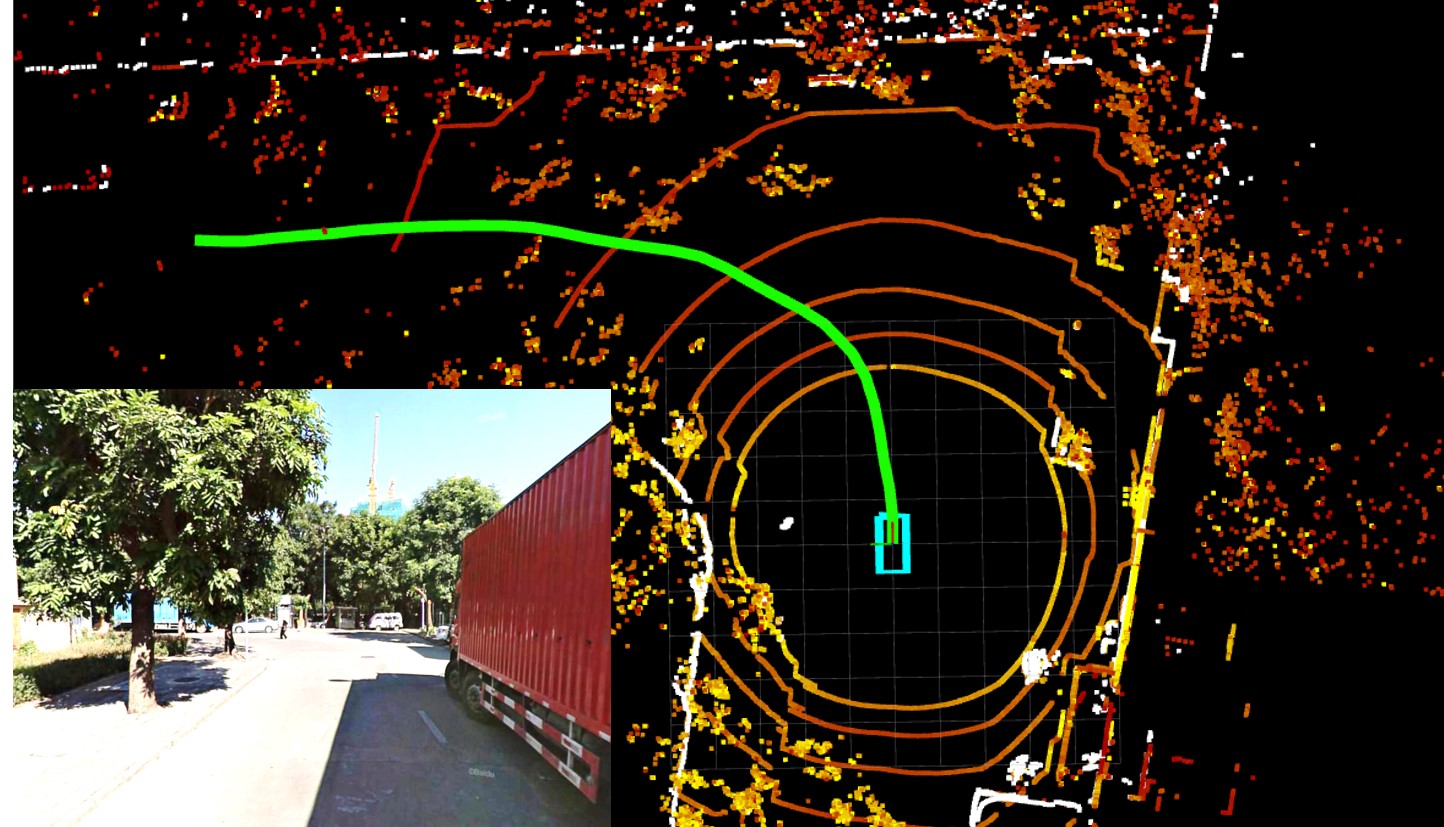}
    }
     \subfigure[Overtake]{\includegraphics[width=0.55\columnwidth]{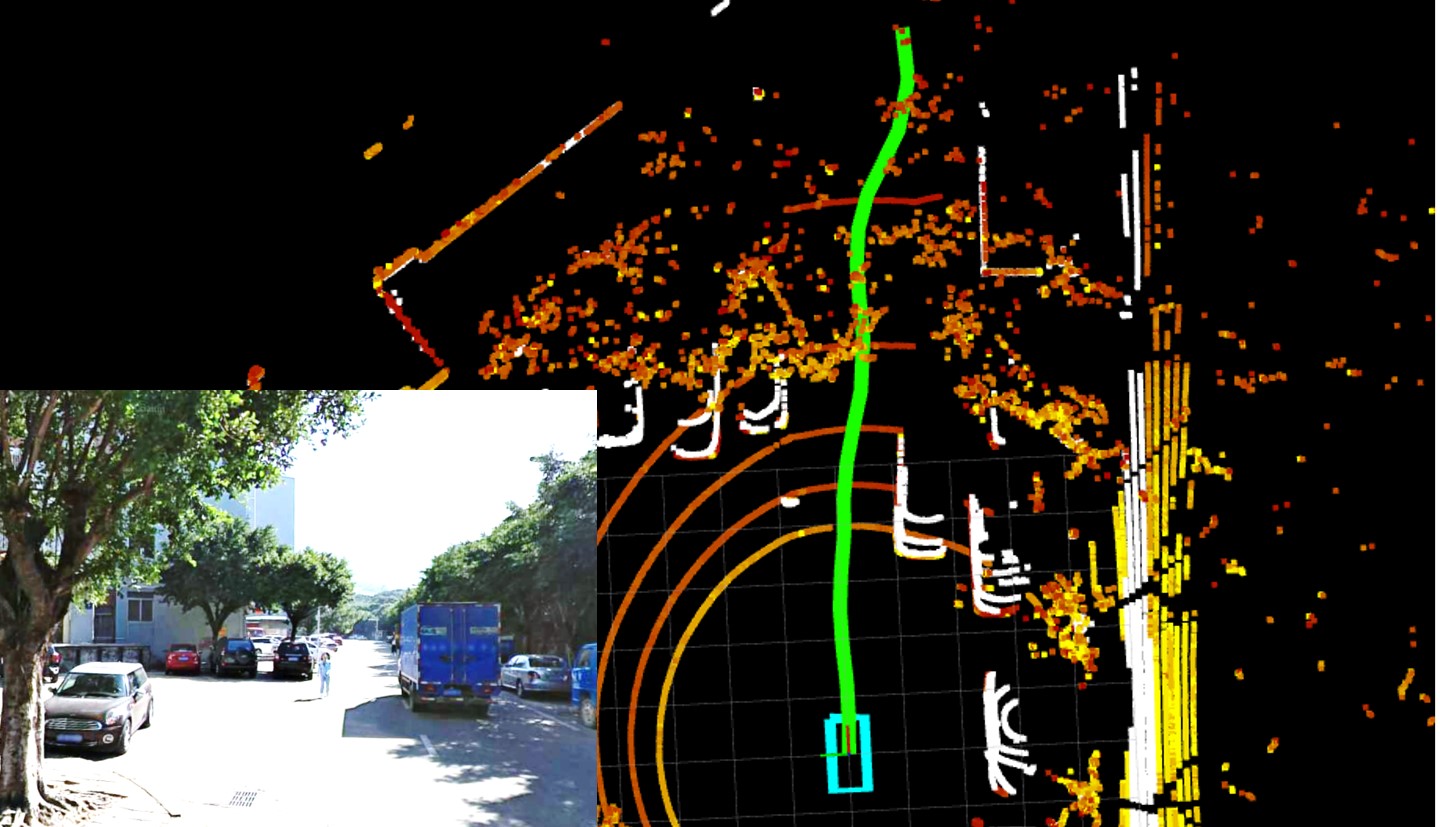}
    } 
  \caption{The extracted different driving scenarios shown with a top-down view. The  yellow points with varied colors represent the lidar scan points of different intensities and the white points represent the filtered laser points for collision checking. The green line represents the planned path.  The insets show the corresponding street view from Baidu Maps\protect\footnotemark.\label{fig:planning}}
   \vspace{-0.5em}
\end{figure*}
\subsection{Temporal optimization}

\begin{figure*}[!ht]
   \centering
    {\includegraphics[width=1.8\columnwidth]{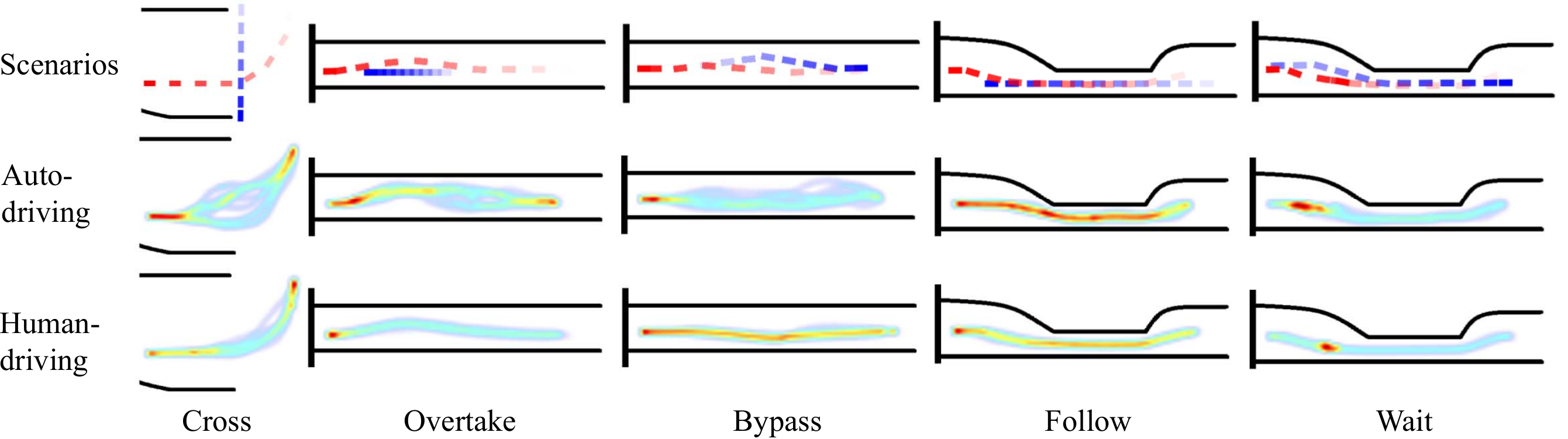} 
  \caption{Heatmaps of the car trace in five different scenarios (better viewed in color). The first row shows the footsteps of the test car (red) and the moving obstacle (blue) for five scenarios. The transparencies increase with time. The second and third row show the heat maps of auto-driving and human-driving trajectories with the data collected with the same interval. The warmer the color, the slower the speed. Video available at: https://youtu.be/Xa9KxVnnyZg.    
  \label{fig:heatmap}} }
  \vspace{-1em}
\end{figure*}
 

\begin{figure}[!htb]
  \centering
	{\includegraphics[width=0.8\columnwidth]{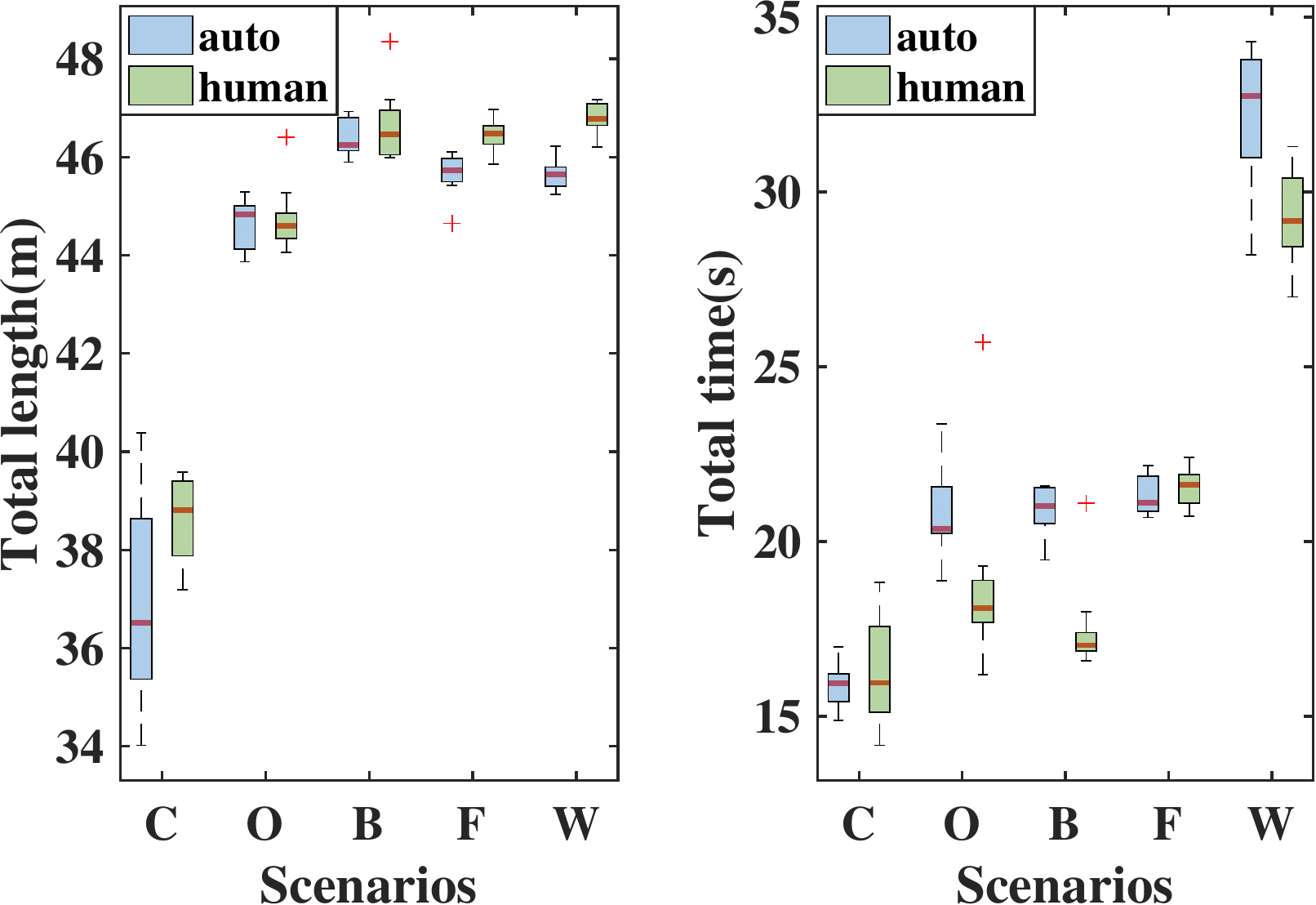} 
  \caption{Total time and total length of trajectories for the auto-driving and human-driven cars in different scenarios (C:Cross O:Overtake B:Bypass F:Follow W:Wait)
  \label{fig:simu-boxplot}} }
\end{figure}

%

For temporal optimization, a critical threat is the dynamic moving obstacles. As shown in Fig.\ref{fig:heatmap}, to test the optimizer in the time domain, we design simulations of five common scenarios, including crossing, overtaking, bypassing, following and waiting. These scenarios are shown in the first row of Fig.\ref{fig:heatmap}, with red boxes representing the test car, and blue representing the moving obstacle. Poses with a larger timestamp have higher transparency. For each scenario, the data of 10 auto-driving tests and 10 human driving tests are collected.   

For qualitative analysis, heat maps showing the distribution of all footsteps along the trajectory are presented. 
The warmer the color, the slower the car moves. 
The values of the pixels are normalized inside a heat map; therefore, the colors of the pixels are only meaningful in a relative sense. 
In the ``Cross" scenario, the red car tries to turn left. Because it cannot cross in front of the blue car in advance even at its maximum speed, it moves slowly and speeds up after the blue car crosses. 
As there is much space for the turning, the auto-driving car goes through different trajectories, while the human-driven car insists on a similar compliant route according to the driving experience in real world.  
In the ``Overtake" scenario, the red car tries to pass a slow car (in blue), while in the ``Bypass" scenario, the red car passes the blue car moving in the opposite direction. Similarly, the trajectories of the auto-driving car vary while the trajectories of human-driven car show consistent patterns. The auto-driving car is not programmed to obey the traffic conventions. More specifically, 
the lane changes to the right are performed at different locations in the ``Overtake" scenario; 
in the ``Bypass" scenario, though it has a tendency to travel to the right due to the motion of the other car, it moves more freely as there are no rigid confinements except the black line. In the ``Follow" scenario, the red car has to travel through a narrow passage with the blue car moving in front, so it has to follow the blue car as there is no room to overtake it. In the ``Wait" scenario, the blue car enters the narrow passage first, and the red car has to wait for the blue car to exit, otherwise the two cars will collide in the passage. From the heat map, we can see that the speeds of the red cars are uniform along the way for the ``Follow" scenario, and both the auto-driving and human-driving cars accelerate after passing the passage area. In the ``Wait" scenario, both the auto-driving and human-driving cars wait at the entrance of the passage. However, the auto-driving car is more conservative and it waits farther away for the entrance to the narrow passage.

\footnotetext{http://quanjing.baidu.com/}
For quantitative analysis, the total time and total length of the trajectories in different scenarios are analyzed. As shown in Fig. \ref{fig:simu-boxplot}, the planned trajectories of the auto-driving cars have a smaller or comparative total length, which can be attributed to the free driving style, especially for the ``Cross" scenario. For the time spent along the way, though the auto-driving car drives through a shorter path, it shows comparative performance in only the ``Cross" and ``Following" scenarios. In other cases, the auto-driving car takes a conservative strategy and takes more time than the human-driving car. Though the time is longer, it is acceptable for a safe trip. 

\subsection{Real test}
To prove the feasibility of the algorithm in the real environment, we build a complex environment and test it with the golf cart (shown in Fig.\ref{fig:function}) equipped with a Velodyne VLP-16 lidar. As other road participants will also contribute to the success of a navigation trial, it is difficult to evaluate the effect of the algorithm in the real environment. In the test, pedestrians move around and the car is noticed or unnoticed, but in either they are acting aggressively. The speed of the car is limited to 1 m/s for safety reasons. The result of the test is shown in Fig.\ref{fig:real-scene}. As no preliminary knowledge of the environment is provided, the path is planned based on current observation and replanning will be conducted if new observations show that the current path is not feasible. For the first row of figures, the golf cart meets a pedestrian who wants to cross the road. As the car cannot travel through safely in advance, it slows down to let the pedestrian pass by. For the second row, two pedestrians travel in the opposite direction. Because the goal is near, there is no other feasible path. So the golf cart waits for the pedestrian to leave and then reaches the end.  
\begin{figure*}[!ht]
   \centering
    {\includegraphics[width=1.6\columnwidth]{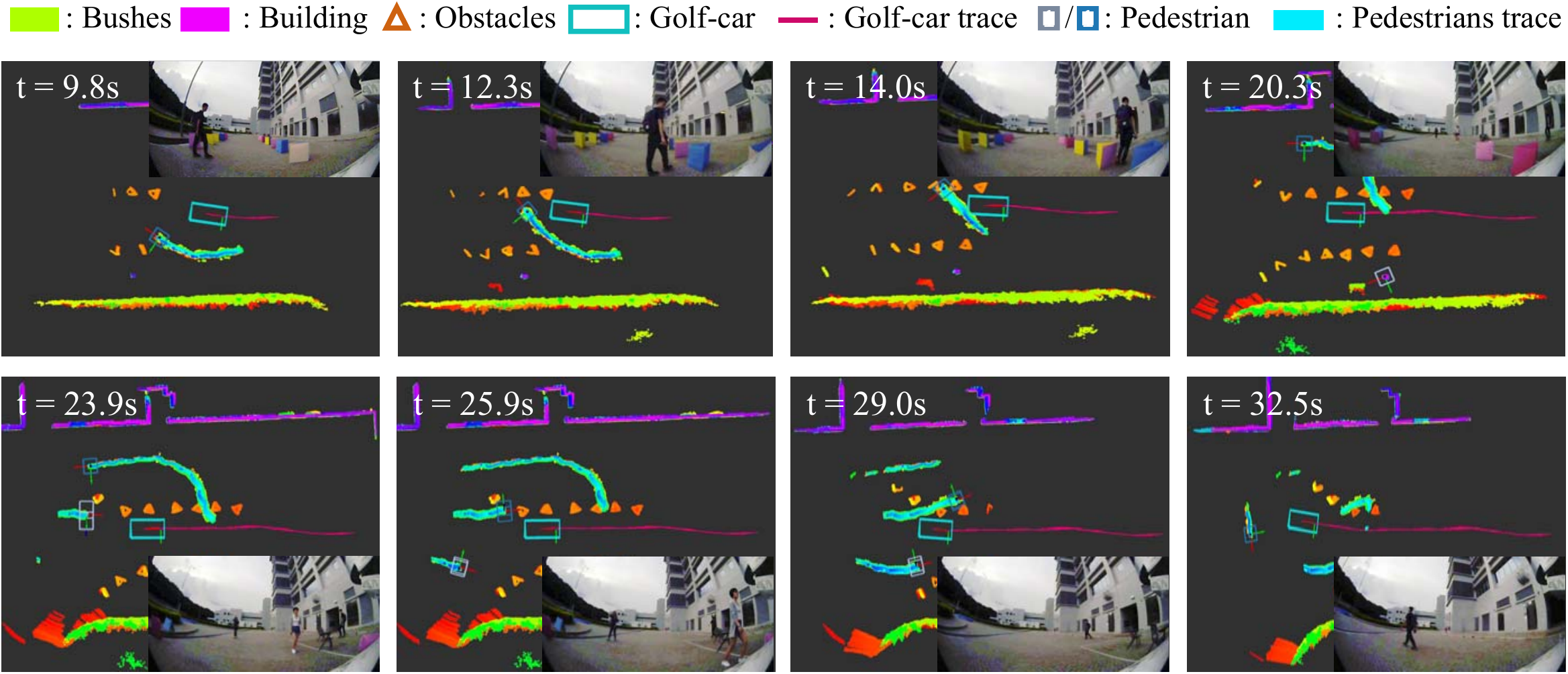}
  \caption{Golf cart navigation in the complex real environment. The cyan box shows the current pose of the golf cart, the gray or blue box shows the current poses of the pedestrians and the blue trails behind them show the tracks. The insets show the front view of the golf cart.
  \label{fig:real-scene}} }
\end{figure*}
\section{CONCLUSIONS}
In this paper, a hierarchical planning approach was proposed to solve the planning and obstacle avoidance problems in dynamic and cluttered environments. A modified bi-RRT planner was proposed to satisfy both the kinematic constraints of a car and the time and flexibility requirements in a dynamic and cluttered environment. Temporal optimization taking advantage of the idea of SI was executed to handle the dynamic environment. 
The planner was evaluated by both simulation and in real tests, and has shown good performance. When compared with human drivers, it showed similar moving patterns handling the narrow passage. Though it planed shorter path, it behaved less agile than human drivers in many scenarios, which can be attributed to the simple modelling of other moving objects.






\
\
\bibliographystyle{IEEEtran}
\bibliography{navibib}
\end{document}